%% file: 0_main.tex
\documentclass[runningheads]{llncs}
\usepackage[T1]{fontenc}
\usepackage{graphicx,verbatim}
\usepackage{amsmath}
\usepackage[T1]{fontenc}
\usepackage{graphicx}
\usepackage{diagbox}
\usepackage{amsfonts}
\usepackage{multirow}
\usepackage{enumitem}
\usepackage{hyperref}
\usepackage{siunitx}
\usepackage{bbm}
\usepackage{pifont}   % For \ding{55} (cross symbol)

\usepackage[capitalise]{cleveref}

\usepackage{xcolor,soul}
\newcommand{\thickhline}{\noalign{\hrule height 1pt}}
\usepackage{float}
\usepackage{color}%, colortbl}
\usepackage{amssymb}  % For symbols like \checkmark

\usepackage{graphicx}
\usepackage{pifont}   % For \ding{55} (cross symbol)
\usepackage{color, colortbl}
\definecolor{Gray}{gray}{0.88}
\definecolor{LightCyan}{rgb}{0.8,1,1}
\definecolor{LightGreen}{rgb}{0.0, 0.5, 0.0}
\definecolor{LightRed}{rgb}{1, 0.7, 0.7}
\definecolor{mydarkgreen}{RGB}{0,160,0}
\usepackage{tabularx} % in preamble
\usepackage{ragged2e}
\newcolumntype{Y}{>{\RaggedRight\arraybackslash}X}
\usepackage{makecell}
\usepackage{array}
\usepackage{multirow}
\usepackage{graphicx}

\newcolumntype{P}[1]{>{\raggedright\arraybackslash}p{#1}}

\begin{document}
\title{Footprint-Guided Exemplar-Free Continual Histopathology Report Generation}
%\title{FG-HRG-Net: Footprint-Guided Exemplar-Free Continual Histopathology Report Generation from Gigapixel Whole Slide Images }

% CHRG-Net: Exemplar-free Continual Histopathology Report Generation from Gigapixel Whole Slide Images
% Generative Latent Replay with Domain Footprints for Continual WSI Report Generation
% FG-HRG-Net: Footprint-Guided Latent Replay for Exemplar-Free Continual Histopathology Report Generation from Gigapixel Whole Slide Images
% CHRG-Net: Footprint-Based Exemplar-Free Continual Histopathology Report Generation
% CHRG-Net: Domain-Footprint Replay for Exemplar-Free Continual Histopathology Report Generation

\titlerunning{Continual WSI Report Generation}

\begin{comment}  %% Removed for anonymized MICCAI 2025 submission
\author{First Author\inst{1}\orcidID{0000-1111-2222-3333} \and
Second Author\inst{2,3}\orcidID{1111-2222-3333-4444} \and
Third Author\inst{3}\orcidID{2222--3333-4444-5555}}
\institute{Princeton University, Princeton NJ 08544, USA \and
Springer Heidelberg, Tiergartenstr. 17, 69121 Heidelberg, Germany
\email{lncs@springer.com}\\
\url{http://www.springer.com/gp/computer-science/lncs} \and
ABC Institute, Rupert-Karls-University Heidelberg, Heidelberg, Germany\\
\email{\{abc,lncs\}@uni-heidelberg.de}}

\end{comment}
\author{Pratibha Kumari\inst{1}\textsuperscript{$@$}\and%index{Kumari, Pratibha}
Daniel Reisenb\"uchler\inst{1}\and %index{Reisenbüchler, Daniel}
Afshin Bozorgpour\inst{1}\and %index{Bozorgpour, Afshin}
yousef Sadegheih\inst{1} \and %index{Sadegheih, yousef}
Priyankar Choudhary\inst{2} \and %index{Choudhary, Priyankar}
Dorit Merhof\inst{1} %index{Merhof, Dorit}
}
\authorrunning{Kumari et al.}
\institute{Faculty of Informatics and Data Science, University of Regensburg, Germany 
\and
Indian Institute of Information Technology Bhopal, India
\\
\textsuperscript{$@$}Correspondence (Pratibha.Kumari@ur.de)
} 
    
\maketitle     
\begin{abstract}
Rapid progress in vision-language modeling has enabled pathology report generation from gigapixel whole-slide images, but most approaches assume static training with simultaneous access to all data. In clinical deployment, however, new organs, institutions, and reporting conventions emerge over time, and sequential fine-tuning can cause catastrophic forgetting. We introduce an exemplar-free continual learning framework for WSI-to-report generation that avoids storing raw slides or patch exemplars. The core idea is a compact domain footprint built in a frozen patch-embedding space: a small codebook of representative morphology tokens together with slide-level co-occurrence summaries and lightweight patch-count priors. These footprints support generative replay by synthesizing pseudo-WSI representations that reflect domain-specific morphological mixtures, while a teacher snapshot provides pseudo-reports to supervise the updated model without retaining past data. To address shifting reporting conventions, we distill domain-specific linguistic characteristics into a compact style descriptor and use it to steer generation. At inference, the model identifies the most compatible descriptor directly from the slide signal, enabling domain-agnostic setup without requiring explicit domain identifiers. 
Evaluated across multiple public continual learning benchmarks, our approach outperforms exemplar-free and limited-buffer rehearsal baselines, highlighting footprint-based generative replay as a practical solution for deployment in evolving clinical settings.

% Authors must provide keywords and are not allowed to remove this Keyword section.

\end{abstract}

\input{1_intro}

\input{2_method_codebook_style}

\input{3_Experiments}

\input{6_ablation}

\input{7_Conclusions}

\bibliographystyle{splncs04}
\bibliography{0_main}

\end{document}

%% file: 1_intro.tex
%%%%%%%%%%%%%%%%%%%%%%%%%%%%%%%%%%%%%%%%%%%%%%%%%%%%%%%%%%%%%%%%%%%%%%%%%%%%%%
\section{Introduction}

Gigapixel whole slide images (WSIs) contain rich morphological information that underlies diagnostic and prognostic decisions in surgical pathology. In routine workflows, this information is summarized in free-text reports that describe salient findings and frequently follow semi-structured conventions (for example, diagnosis, grade, stage, and margin status). Automating report generation from WSIs is therefore an important step toward reducing documentation burden and enabling more consistent reporting. Recent vision-language models have begun to translate WSIs into coherent diagnostic narratives by combining slide encoders with large language models (LLMs), enabling multi-scale reasoning over billions of pixels and producing clinically meaningful text~\cite{chen2024wsicaption,tran2025generating,tran2025generating}.

Despite this progress, most WSI report-generation systems are developed in a static training regime that assumes simultaneous access to multiple datasets spanning different organs, institutions, scanners, and patient populations. In practice, deployment is inherently continual, with new datasets (domains) arriving over time. Repeated cumulative retraining on all observed data can become computationally costly and may be infeasible when prior data cannot be retained due to storage and governance constraints~\cite{sadegheih2025modality}. Na\"{\i}ve fine-tuning on incoming data can lead to catastrophic forgetting (CF), where performance degrades on previously learned domains~\cite{sadegheih2026towards,kumari2025domain,kumari2025attention}. This challenge is pronounced in report generation when domain shifts affect both the input distribution and the report language, requiring CL methods that preserve input-output drift over time.

A widely used mechanism to reduce forgetting is rehearsal, where past examples are interleaved with current-domain training~\cite{rolnick2019experience,pellegrini2020latent,bhatt2024characterizing}. In computational pathology, however, rehearsal is challenging because WSIs are naturally represented as variable-length sets of patch embeddings, making storage costly even when retaining only intermediate features rather than raw pixels; in addition, long-term retention of patient data may be restricted in some settings~\cite{Kum_Continual_MICCAI2024}. Beyond exemplar replay, prompt-based CL for LLMs learns task-specific soft prompts while keeping the backbone frozen, but it typically assumes task or domain identity at inference to apply the appropriate prompt sequence~\cite{razdaibiedina2023progressive}. Relatedly, recent work on continual radiology report generation uses parameter-efficient tuning, learning a small set of domain-specific parameters while freezing the LLM backbone~\cite{sun2024continually}; however, such strategies still maintain domain-specific components whose overhead grows as domains accumulate. In computational pathology, domain identifiers or metadata may be missing, inconsistent across sites, or insufficient to reflect shifts in reporting conventions, which can reduce robustness at deployment. Together, these considerations motivate a CL framework for WSI report generation that is storage-efficient, avoids retaining raw exemplars, and supports domain-agnostic inference.

In this work, we introduce a footprint-based generative replay framework for continual WSI report generation. Building on a HistoGPT-style vision-language generator~\cite{tran2025generating}, we address the core constraint of continual deployment: retaining prior-domain competence without keeping WSIs or large feature archives. Our key idea is to compress each domain into a compact domain footprint in a frozen embedding space, combining a representative codebook with lightweight slide-level composition statistics. These footprints enable exemplar-free rehearsal by synthesizing pseudo-WSIs, while pseudo reports are obtained from an immediate teacher to provide consistent supervision under the same next-token objective as current-domain training. 
Since latent generation operates in this frozen space, it avoids generator drift across domains.
To further cope with shifts in reporting conventions, we incorporate per-domain \textit{report-style prototype} and condition the LLM through a lightweight style-prefix. At inference, we infer the most compatible footprint directly from slide content, enabling domain-agnostic generation. We evaluate our approach under domain-incremental scenarios, measuring report quality and retention across domains.

\paragraph{Contributions.}
We propose a CL framework for WSI report generation that
(i) represents each domain with compact visual footprints in a frozen embedding space, enabling rehearsal without storing raw WSIs or features,
(ii) enables exemplar-free replay via footprint-based pseudo-WSIs and immediate-teacher pseudo reports, with performance comparable to exemplar replay, and
(iii) supports domain-agnostic, style-conditioned generation using per-domain report-style prototypes.

%% file: 2_method_codebook_style.tex
%%%%%%%%%%%%%%%%%%%%%%%%%%%
\begin{figure}[!t]
\centering
\includegraphics[scale=0.6]{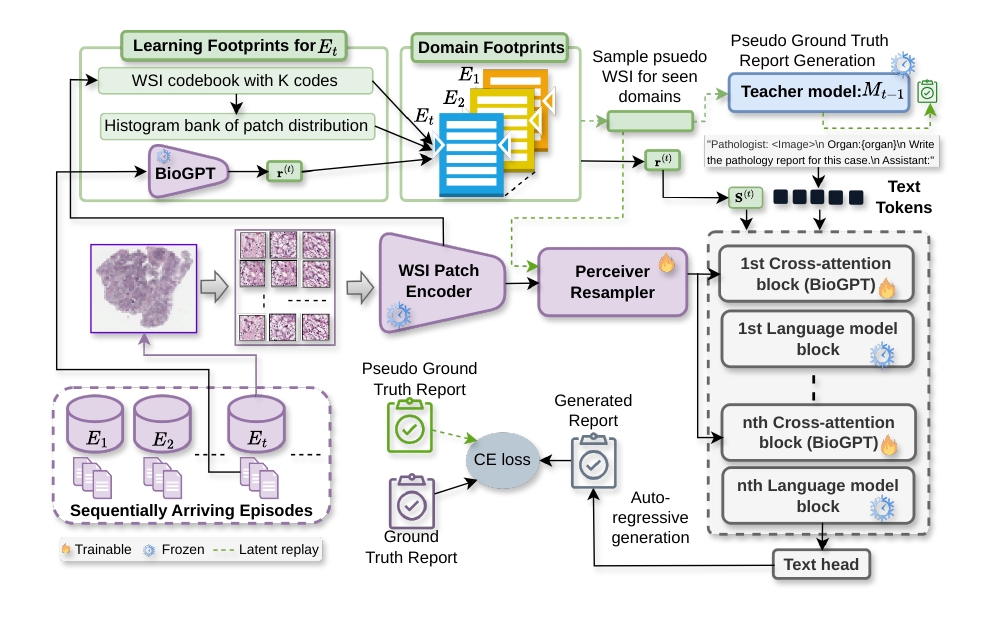}
\caption{Overview of the proposed continual WSI report generation framework. }
\label{fig:placeholder}
\end{figure}
%%%%%%%%%%%%%%%%%%%%%%%%%%%

\section{Method}
We study continual WSI report generation in a domain-incremental setting. Training proceeds over domains, also termed as episodes $\{E_t\}_{t=1}^{T}$, where at episode $t$ the model has access only to the current dataset $\mathcal{D}_t=\{(\mathbf{X},y)\}$ and cannot revisit samples from earlier episodes $\{\mathcal{D}_1,\dots,\mathcal{D}_{t-1}\}$. The goal is to learn each new domain while maintaining performance on previously learned domains. Our framework (Figure~\ref{fig:placeholder}) has three components: (i) compact per-domain footprints that summarize morphology in a frozen embedding space, (ii) footprint-based replay that synthesizes pseudo-WSIs and pseudo reports to rehearse prior domains, and (iii) a per-domain report-style prototype injected into the language model to handle shifts in reporting conventions. The proposed CL strategy is model-agnostic and can be attached to any WSI-to-report generator that consumes patch embeddings; we instantiate it with a HistoGPT architecture~\cite{tran2025generating}.
 
\subsection{Report Generation for Giga-pixel Histopathology Images}
A WSI $x$ is represented as a set of patch embeddings $\mathbf{X}=\{\mathbf{x}_i\}_{i=1}^{N}$, where $\mathbf{x}_i\in\mathbb{R}^{D}$ and $N$ varies with tissue size and tiling. Patch embeddings are extracted with a frozen visual encoder. As in prior work, a Perceiver Resampler~\cite{jaegle2021perceiver} maps the variable-length set $\mathbf{X}$ to a fixed-length latent sequence $\mathbf{Z}\in\mathbb{R}^{L_v\times D_{\mathrm{LM}}}$, which serves as visual context for language generation. An autoregressive language model then generates report tokens $y=(y_1,\dots,y_L)$ conditioned on $\mathbf{Z}$ via cross-attention, using an organ-specific prompt~\cite{tran2025generating}. Training uses teacher forcing with next-token cross-entropy. Let $q$ denote prompt tokens and $y$ the report tokens. We minimize cross-entropy over the concatenated sequence $[q;y]$ while masking prompt positions, so the loss is computed only on report tokens.

\subsection{Report Generation with Report-style Prototype}\label{sec:report-style}
Reporting conventions can vary across domains (e.g., template, verbosity and phrasing). To provide a constant-size style cue, we compute a single report-style prototype for each domain $t$. Domain reports are encoded with a frozen text encoder, and the prototype $\mathbf{r}^{(t)}\in\mathbb{R}^{D_{text}}$ is obtained by mean-pooling token embeddings followed by $\ell_2$ normalization. The prototype is computed once when a domain is first observed and is used to condition both real samples from $\mathcal{D}_t$ and future pseudo replay samples associated with domain $t$.

We condition the language model via style-prefix tokens. The tokenizer is extended with $M$ dedicated style tokens $s=(s_1,\dots,s_M)$ that are prepended to the textual input. A learned linear projection maps the prototype to prefix embeddings in the LM space, $\mathbf{S}^{(t)}=\phi(\mathbf{r}^{(t)})\in\mathbb{R}^{M\times D_{\mathrm{LM}}}$,
where $\phi:\mathbb{R}^{D_{text}}\rightarrow\mathbb{R}^{M\times D_{\mathrm{LM}}}$. During the forward pass, the input embeddings at the style-token positions are replaced by $\mathbf{S}^{(t)}$, while all other tokens use the standard embedding table. The resulting sequence $[s;q;y]$ is then processed by the autoregressive decoder with cross-attention to the visual latents $\mathbf{Z}$. Style-token positions are excluded from the loss by masking their labels, so training remains next-token cross-entropy on report tokens.

\subsection{Domain Footprints}
To support rehearsal without storing WSIs or per-slide patch features, we summarize each domain $t$ with a compact domain footprint computed in the frozen patch-embedding space. For footprint construction, we subsample a bounded number of WSIs and, per WSI, a bounded number of patch embeddings, yielding a filtered set $\mathbf{X}^{\star}$. We learn a domain codebook by $k$-means clustering,
$\mathbf{C}^{(t)}=\{\mathbf{c}^{(t)}_k\}_{k=1}^{K}$ with $\mathbf{c}^{(t)}_k\in\mathbb{R}^{D}$.
Each patch embedding $\mathbf{x}_i \in \mathbf{X}^{\star}$ is quantized to its nearest codeword,
$
k_i=\arg\min_{k}\|\mathbf{x}_i-\mathbf{c}^{(t)}_k\|_2^2,
$
turning a WSI into a multiset of discrete code indices with fixed storage per domain.
To retain slide-level composition, we compute for each training slide a normalized histogram $\mathbf{h}\in\mathbb{R}^{K}$ over code assignments,
$
h_k=\frac{1}{N}\sum_{i=1}^{N}\mathbbm{1}[k_i=k],
$
and store a small histogram bank $\mathcal{H}^{(t)}$. We additionally store patch-count statistics $(\mu_N^{(t)},\sigma_N^{(t)})$ to reproduce realistic slide sizes during replay. The domain footprint is
$
\mathcal{F}^{(t)}=\{\mathbf{C}^{(t)},\mathcal{H}^{(t)},\mu_N^{(t)},\sigma_N^{(t)},\mathbf{r}^{(t)}\},
$
where $\mathbf{r}^{(t)}$ denotes the report-style prototype (Sec.~\ref{sec:report-style}).

\subsection{Generative Replay with Pseudo Reports}
At episode $t$, training interleaves supervised updates on $\mathcal{D}_t$ with replay sampled from footprints of past domains. To synthesize a pseudo WSI for a previous domain $j<t$, we sample a patch count $N\sim\mathcal{N}(\mu_N^{(j)},(\sigma_N^{(j)})^2)$ and draw a histogram $\tilde{\mathbf{h}}\in\mathcal{H}^{(j)}$. We then sample code indices $k_1,\dots,k_N$ from the categorical distribution defined by $\tilde{\mathbf{h}}$ and map them to codewords, yielding a synthetic patch set
$\hat{\mathbf{X}}^{(j)}=\{\mathbf{c}^{(j)}_{k_i}\}_{i=1}^{N}$.
Optionally, we add small Gaussian noise to the sampled codewords to increase diversity.

Because $\hat{\mathbf{X}}^{(j)}$ has no ground-truth report, we generate a pseudo target using an \emph{immediate teacher}, i.e., a frozen snapshot of the model at the end of episode $t{-}1$. Conditioned on $\hat{\mathbf{X}}^{(j)}$ and the prompt, the teacher produces a pseudo report $\tilde{y}$, and the current model is trained on $(\hat{\mathbf{X}}^{(j)},\tilde{y})$ with the same next-token cross-entropy as for real data. Both real and replay samples are conditioned using the report-style prototype of their domain. The $t^{th}$ episode objective is
\begin{equation}
\mathcal{L}_t \;=\;
\mathbb{E}_{(\mathbf{X},y)\sim \mathcal{D}_t}\!\left[\mathcal{L}_{\mathrm{CE}}(y \mid \mathbf{X}, \mathbf{r}^{(t)})\right]
\;+\; \lambda\, \mathbb{E}_{(\hat{\mathbf{X}},\tilde{y},j)\sim \mathcal{R}_{<t}}\!\left[\mathcal{L}_{\mathrm{CE}}(\tilde{y} \mid \hat{\mathbf{X}}, \mathbf{r}^{(j)})\right].
\end{equation}
with $\mathcal{R}_{<t}$ denoting pseudo tuples $(\hat{\mathbf{X}},\tilde{y},j)$ from domains $j<t$, and $\mathcal{L}_{\mathrm{CE}}(\cdot;\mathbf{r})$ the next-token cross-entropy under style conditioning.

\subsection{Domain-Agnostic Inference}
At test time, explicit domain identifiers may be unavailable. We therefore select the most compatible footprint directly from the slide content and use its report-style prototype for conditioning. Given a test slide embedding set $\mathbf{X}=\{\mathbf{x}_i\}_{i=1}^{N}$, we score each domain $t$ by its codebook reconstruction error in the frozen embedding space,
$\mathrm{QE}(t;\mathbf{X})=\frac{1}{N}\sum_{i=1}^{N}\min_{k\in\{1,\dots,K\}}\|\mathbf{x}_i-\mathbf{c}^{(t)}_k\|_2^2$,
and choose $\hat{t}=\arg\min_t\,\mathrm{QE}(t;\mathbf{X})$. Report generation is then conditioned using the selected prototype $\mathbf{r}^{(\hat{t})}$ via the style-prefix mechanism, enabling domain-agnostic generation without relying on dataset metadata.

%% file: 3_Experiments.tex
\section{Experiment and Results}\label{sec:exp}

\subsection{Experimental Setup}
\noindent\textbf{Datasets.}
We evaluate WSI report generation on two public datasets. REG2025~\cite{reg2025_dataset} provides WSI-report pairs from 7 organs, arranged as an organ-shift (OS) sequence with one organ per episode. PathText~\cite{chen2024wsicaption}, curated from TCGA~\cite{TCGA}, is re-partitioned into an OS benchmark with one organ per episode; we keep 6 organs with at least 500 pairs each. We also construct 2 hybrid 6-episode streams by considering common and disjoint organs across REG2025 and PathText, yielding heterogeneous shifts (HS), including larger report-style differences. Dataset statistics, experiment names, and episode composition are provided in Table~\ref{tab:datasets_detail}.

\noindent\textbf{Baselines.}
We compare against representative CL methods. Rehearsal-based: ER~\cite{rolnick2019experience} and DER~\cite{buzzega2020dark}. Regularization-based: EWC~\cite{kirkpatrick2017overcoming}, SI~\cite{zenke2017continual}, and LwF~\cite{li2017learning}. We additionally include ProgPrompt~\cite{razdaibiedina2023progressive}, which grows domain-specific soft prompts and selects the appropriate prompt using the domain identity. We also include the continual radiology report generation method, CMRG-LLM~\cite{sun2024continually}, which demands domain-specific sub-networks; we report results without their label-based alignment loss because disease labels are not available in our WSI data. Sequential fine-tuning (Na\"ive) and From-Scratch Training (FST) serve as lower-bound (LB), whereas Joint and Cumulative provide upper-bound (UB) references~\cite{Kum_Continual_MICCAI2024}.

\input{4_datasets}

\noindent\textbf{Implementation Details and Evaluation Metrics.}
WSIs are tiled into patches with CLAM~\cite{lu2021data} and encoded using frozen CONCH visual encoder~\cite{conch}. All methods use the same HistoGPT backbone and prompt format. Models are trained for 10 epochs using a learning rate of $\num{5e-5}$. ER/DER use a fixed exemplar budget of 10-50 samples, and regularization baselines follow standard settings from prior CL work~\cite{Kum_Continual_MICCAI2024}. (\#) token for ProgPrompt is set to 50. For our method, we set $M$=4, $K$=64, $\mathcal{H}^{(t)}$=50 per domain for footprint construction, and generate 20 pseudo WSI-report pairs per domain. The report-style prototypes are computed from frozen BioGPT~\cite{luo2022biogpt}.
We follow REG2025 evaluation protocol~\cite{reg2025_dataset} and compute the composite ranking score\footnote{\textit{Ranking\_score}: $0.15(\mathrm{ROUGE}+\mathrm{BLEU4})+0.4\,\mathrm{key\_score}+0.3\,\mathrm{emb\_score}$.}. For a sequence of $T$ episodes, we evaluate after each episode to obtain a $T\times T$ matrix of composite scores, from which we compute standard CL metrics: AVG, ILM, and BWT~\cite{lopez2017gradient,diaz2018don,kumari2023continual}.
%%%%%%%%%%%%%%%%%%%%%%%%%%%%%%%%%%%%%%%%%

%%%%%%%%%%%%%%%%%%%%%%%%%%%%%%%%%%%%%%%%%%%%%%%%%%%%%%%
\subsection{Results}

\input{4_Table_main}

Table~\ref{tab:comp_cl_all} lists performance on four experiments (OS-R, OS-P, HS-C, HS-D) using AVG, ILM, and BWT computed from \textit{Ranking\_score}. For OS-R (Reg2025 organ shift), sequential fine-tuning on HistoGPT (Na\"ive and FST) suffer severe CF, evidenced by strongly negative BWT and low average performance (AVG and ILM). Joint and Cumulative provide upper bounds but require simultaneous access to all datasets, along with substantial storage and computation. 
Rehearsal methods are effective when the exemplar budget ($\mathcal{B}$) is large (DER ($\mathcal{B}$=50): 0.6620/0.7144/-0.0883) but, the performance degrades notably with smaller $\mathcal{B}$ (DER ($\mathcal{B}$=10): 0.6299/0.6092/-0.2246). 
Among exemplar-free baselines, regularization methods EWC, SI and LwF largely collapse to Na\"ive behavior (e.g., OS-R, ILM remains $\approx$0.40), suggesting that constraining parameter drift alone is insufficient for this report generation application. ProgPrompt yields zero forgetting by construction since the backbone is frozen and only domain-specific prompts are learned then frozen, resulting in BWT=0 on all sequences; however, restricted plasticity reduces average performance (OS-R, AVG/ILM: 0.6584/0.6680) and it requires domain identity at inference to activate the appropriate prompt. CMRG-LLM performs poorly in our WSI setup, consistent with the absence of explicit disease labels and limited vision-text interaction when the LLM is fully frozen. Our exemplar-free method achieves AVG/ILM/BWT: 0.6961/0.7319/-0.0686, substantially outperforming rehearsal methods having small buffer ($\mathcal{B}$=10, 20) and matching performance with rehearsal methods having large buffer. Thus, our method, which operates without storing exemplars or requiring domain identity at inference, demonstrates that footprint-based generative replay serves as an effective rehearsal alternative.

For OS-P (organ shift on PathText), absolute performance is consistently lower than OS-R across methods. This sequence is more challenging because reports frequently include clinical context that is only weakly grounded in slide morphology, which limits the maximum attainable performance even for non-CL UB. This trend is reflected in the smaller improvement from Na\"ive to Cumulative method on OS-P (ILM: 6.25\%) relative to OS-R/HS-C/HS-D (89.59/87.67/89.91)\%. 
Within this constrained regime, our method remains a practical alternative that requires neither domain identifiers nor real exemplars.

For HS-C/HS-D, performance degrades across methods due to the presence of heterogeneous shift that combine organ changes with variations in report style. In this setting, our method again outperforms other exemplar-free approaches while avoiding domain-identifiers. 
Rehearsal methods exhibit degradation with small $\mathcal{B}$ (e.g., DER drops by 7.3/3.9/87.9\% with $\mathcal{B}$=50$\rightarrow$20 on HS-C).
Compared to the best low-buffer result (blue), our method improves HS-C (AVG/ILM/BWT) by (3.5/1.8/7.1)\% and HS-D by (6.5/5.6/41.6)\%. Thus, our method benefits from report-style prototype conditioning, which improves both average performance and retention under HS.

Table~\ref{tab:qualitative} presents qualitative results for two Bladder ($E_1$) cases evaluated after training up to Stomach ($E_T$) in OS-R experiment. (Na\"ive) shows pronounced cross-organ drift, generating Stomach-specific reports (e.g., MALT lymphoma or gastrointestinal stromal tumor) instead of Bladder findings. DER mitigates this drift but still introduces inconsistent clinical details. In contrast, our method preserves organ context and produces Bladder-consistent reports across both cases, demonstrating improved performance retention without storing past slides.

\input{4_TABLE_QUALITATIVE}

%% file: 4_datasets.tex
\begin{table}[!t]
\centering
\caption{Details of CL datasets and train-test used in various experiments.}
\label{tab:datasets_detail}
\resizebox{\textwidth}{!}{%
\begin{tabular}{c|c|c|c|c}
\thickhline

{\bf Source} &
{\bf Exp. } &
  {\bf (\#) Episodes information} &
  {\bf \#Train } &
  \begin{tabular}[c]{@{}c@{}}{\bf \#Test }\end{tabular} \\ \thickhline

\begin{tabular}[c]{@{}c@{}}REG2025 \end{tabular} &OS-R&
\begin{tabular}[c]{@{}c@{}}(7) Breast$\rightarrow$Bladder$\rightarrow$Cervix$\rightarrow$Colon \\$\rightarrow$Lung$\rightarrow$Prostate$\rightarrow$Stomach \end{tabular} &
\begin{tabular}[c]{@{}c@{}} 1746, 759, 511, 741, \\636, 652, 1307 \end{tabular} &
100 each \\ \hline

\begin{tabular}[c]{@{}c@{}}PathText \end{tabular} &OS-P&
\begin{tabular}[c]{@{}c@{}} (6) breast$\rightarrow$lung$\rightarrow$kidney$\rightarrow$colorectal$\rightarrow$uterus$\rightarrow$thyroid\end{tabular}&
\begin{tabular}[c]{@{}c@{}}948,  820,  774, 485, 451, 403 \end{tabular}&
100 each \\ \hline

\begin{tabular}[c]{@{}c@{}}Common organs\\(REG2025, PathText) \end{tabular} &HS-C&
\begin{tabular}[c]{@{}c@{}} (6) REG2025: (Breast$\rightarrow$Colon$\rightarrow$Lung)$\rightarrow$ \\ PathText: (breast$\rightarrow$colorectal$\rightarrow$ lung)\end{tabular}&
\begin{tabular}[c]{@{}c@{}}1746, 741, 636, \\948, 485,  820 \end{tabular}&
100 each \\ \hline

\begin{tabular}[c]{@{}c@{}}Disjoint organs\\(REG2025, PathText)\end{tabular} &HS-D&
\begin{tabular}[c]{@{}c@{}}(6) REG2025: (Bladder$\rightarrow$Cervix$\rightarrow$Prostate)\\ PathText: (kidney$\rightarrow$uterus$\rightarrow$thyroid) \end{tabular} &
\begin{tabular}[c]{@{}c@{}}  759, 511,  652\\  774, 451, 403 \end{tabular} &
100 each \\ 

\thickhline
  
\end{tabular}
}
\end{table}
%%%%%%%%%%%%%%%%%%%%%%%%%%%%%%%%%%%%%%%%%%%%%%%%%%%%

%% file: 4_Table_main.tex
\begin{table*}[!t]
\centering
\tiny
\caption{Best result in exemplar-based / exemplar-based with low size / exemplar-free categories indicated in \textcolor{red}{red} / \textcolor{blue}{blue} / \textcolor{mydarkgreen}{green}, respectively. \textbf{Bold}: UB, $\mathcal{B}$: exemplar size. }
 \label{tab:comp_cl_all}

\begin{tabular}{c|c|l|ccc|ccc|ccc|ccc}
\thickhline
&\multirow{2}{*}{$\mathcal{B}$}
& \begin{tabular}[c]{@{}c@{}}Exp.$\rightarrow$  \end{tabular}
& \multicolumn{3}{c|}{OS-R }
& \multicolumn{3}{c|}{OS-P }
& \multicolumn{3}{c|}{HS-C  }
& \multicolumn{3}{c}{HS-D  }
\\ \cline{3-15}
 
&&Approach
&  AVG $\uparrow$ &ILM $\uparrow$& BWT $\uparrow$
&  AVG $\uparrow$ &ILM $\uparrow$& BWT $\uparrow$
&  AVG $\uparrow$ &ILM $\uparrow$& BWT $\uparrow$
&  AVG $\uparrow$ &ILM $\uparrow$& BWT $\uparrow$
 \\ \thickhline

\multirow{4}{*}{\rotatebox{90}{Non-CL}} 
&\multirow{4}{*}{-}
&na\"{\i}ve 
& 0.3535 & 0.4065 & -0.5086 
& 0.3428 & 0.3423 & -0.0308  
&  0.2949 & 0.3545 & -0.4239 
&0.2974& 0.3505& -0.4376
\\ 

&
&FST
&0.3502& 0.3942& -0.5026
&0.3389& 0.3414& -0.0303
&0.2910& 0.3524& -0.4230
& 0.2945& 0.3483&-0.4427
\\ \cline{3-15}

&& Joint 
&0.7381 &- &-
&\textbf{0.3734} & -&-
& 0.5687 & -&-
&\textbf{0.5702}&-&-
\\ 
 	
&& Cumulative 
&  \textbf{0.7426} &\textbf{ 0.7707} & \textbf{-0.0312}
& 0.3668 &\textbf{0.3637 }& \textbf{-0.0089} 
& \textbf{0.5692}&\textbf{0.6653}& \textbf{-0.0180}
& 0.5672& \textbf{0.6516}&\textbf{ -0.0420}
\\  \hline

\multirow{12}{*}{\rotatebox{90}{CL}}
&\multirow{6}{*}{\ding{51}}
&\begin{tabular}[c]{@{}c@{}}ER ($\mathcal{B}$=50) \end{tabular} 
&\textcolor{red}{ 0.6714}& 0.7109& -0.0916
& \textcolor{red}{ 0.3549}& 0.3559& \textcolor{red}{-0.0121}
& 0.5251& 0.6141& -0.0596
&0.5308& 0.6289&\textcolor{red}{ -0.0456}\\ 

&&\begin{tabular}[c]{@{}c@{}}DER ($\mathcal{B}$=50) \end{tabular} 
& 0.6620&\textcolor{red}{ 0.7144}&\textcolor{red}{-0.0883}
&  0.3524& \textcolor{red}{0.3594}& -0.0123
&\textcolor{red}{0.5392}& \textcolor{red}{0.6231}& \textcolor{red}{-0.0487}
& \textcolor{red}{0.5480}&\textcolor{red}{ 0.6393}& -0.0483
\\ \cline{3-15}

&&\begin{tabular}[c]{@{}c@{}}ER ($\mathcal{B}$=10) \end{tabular} 
& 0.6100& 0.6099& -0.2186
&0.3461&0.3460& -0.0206
&0.4866&0.5773& -0.1137
&0.4502& 0.5444& -0.1708
\\ 

&&\begin{tabular}[c]{@{}c@{}}ER ($\mathcal{B}$=20) \end{tabular} 
&0.5992 & 0.6431 & -0.1735
&0.3527 & 0.3540 & -0.0139  
& 0.4943 & 0.5952 & -0.0982 
&0.5049& \textcolor{blue}{0.6083}&\textcolor{blue}{-0.0707}
\\

&&\begin{tabular}[c]{@{}c@{}}DER ($\mathcal{B}$=10) \end{tabular} 
& 0.6299 & 0.6092 &-0.2246  
& 0.3493&0.3498& -0.0212
&0.4747& 0.5650& -0.1273
&0.4699&0.5662&-0.1471\\  

&&\begin{tabular}[c]{@{}c@{}}DER($\mathcal{B}$=20)  \end{tabular} 
&\textcolor{blue}{0.6648} & \textcolor{blue}{0.6778} & \textcolor{blue}{-0.1251} 
& \textcolor{blue}{ 0.3570}
&\textcolor{blue}{0.3563} &\textcolor{blue}{-0.0112}
& \textcolor{blue}{0.5000} &\textcolor{blue}{ 0.5989} & \textcolor{blue}{-0.0915 }
&\textcolor{blue}{0.5139}& 0.6032& -0.0761\\
\cline{2-15}

&\multirow{7}{*}{\ding{55}}
& \begin{tabular}[c]{@{}c@{}}SI \end{tabular}
& 0.3211& 0.3823& -0.0012
& 0.3284& 0.3317&-0.0032
& 0.3409& 0.3995& -0.0044
& 0.3416& 0.4037& -0.0092
\\

&&\begin{tabular}[c]{@{}c@{}}LwF \end{tabular} 
& 0.3542 & 0.3998 & -0.4972 
&   0.3475 & 0.3433 & -0.0219  
&0.2949 &0.3615 & -0.4324  
&0.2966& 0.3543& -0.4244
\\

&&\begin{tabular}[c]{@{}c@{}}EWC \end{tabular}
& 0.3567&0.4057& -0.4960
& 0.3425& 0.3442& -0.0313
& 0.2879 & 0.3495 & -0.4242 
& 0.2951& 0.3466& -0.4377
\\

&&\begin{tabular}[c]{@{}c@{}}ProgPrompt \end{tabular}
&0.6584& 0.6680&\textcolor{mydarkgreen}{ 0.0000}
&\textcolor{mydarkgreen}{0.3624}& \textcolor{mydarkgreen}{0.3586}& \textcolor{mydarkgreen}{0.0000}
&0.5131& 0.5750&\textcolor{mydarkgreen}{0.0000}
&0.5265& 0.5939& \textcolor{mydarkgreen}{0.0000}
\\

&&\begin{tabular}[c]{@{}c@{}}CMRG-LLM \end{tabular}
& 0.2998 & 0.3008 & \textcolor{mydarkgreen}{0.0000 }
&  0.2327 & 0.2295 &\textcolor{mydarkgreen}{0.0000 }
&  0.2765& 0.2911& \textcolor{mydarkgreen}{0.0000}
&0.2148& 0.2259& \textcolor{mydarkgreen}{0.0000}
\\

&&\begin{tabular}[c]{@{}c@{}}Our \end{tabular}
& \textcolor{mydarkgreen}{0.6961}&\textcolor{mydarkgreen}{ 0.7319}& -0.0686
&  0.3517& 0.3544&-0.0130
&\textcolor{mydarkgreen}{0.5176}& \textcolor{mydarkgreen}{0.6094}& -0.0850
& \textcolor{mydarkgreen}{0.5473}&\textcolor{mydarkgreen}{0.6424}& -0.0413
\\ \thickhline

\end{tabular}%
\end{table*}
%===============================================================

%% file: 4_TABLE_QUALITATIVE.tex
\begin{table}[!t]
\centering
\tiny

\setlength{\tabcolsep}{4pt}
\renewcommand{\arraystretch}{1.05}

\caption{Qualitative OS-R results. Two Bladder ($E_1$) test cases (GT vs. generated) after training to Stomach ($E_T$). \textcolor{blue}{Blue}: GT-consistent; \textcolor{red}{Red}: mismatches.}
\begin{tabular}{p{0.01cm}|p{0.5cm}|p{10.9cm}}
% \begin{tabular}{p{0.01cm}|p{0.6cm}|p{0.95\linewidth}} 
% \begin{tabular}{c|P{0.09\linewidth}|P{0.8\linewidth}}
\thickhline

\multirow{6}{*}{\rotatebox{90}{Example 1}}&
GT &  Urinary bladder, transurethral resection; No tumor present Note) The specimen includes muscle proper 
 \\\cline{2-3}
 &  Na\"{\i}ve& \textcolor{red}{Stomach, endoscopic biopsy; Extranodal marginal zone B cell lymphoma of MALT type} \\\cline{2-3}
 
&DER $\mathcal{B}$=10
 &\textcolor{blue}{Urinary bladder, transurethral resection; 1. No tumor present} \textcolor{red}{2. Chronic granulomatous inflammation with foreign body reaction} \textcolor{blue}{Note) The specimen includes muscle proper}. \\\cline{2-3}
 
  & DER $\mathcal{B}$=50& \textcolor{blue}{Urinary bladder, transurethral resection;} \textcolor{red}{Chronic nonspecific cystitis}\\\cline{2-3}

  &  Ours&\textcolor{blue}{Urinary bladder, transurethral resection; No tumor present Note) The specimen includes muscle proper.}
  \\\hline

%=====================================%

\multirow{8}{*}{\rotatebox{90}{Example 2}}&GT & Urinary bladder, transurethral resection; Non-invasive papillary urothelial carcinoma, low grade Note) The specimen does not include muscle proper. \\\cline{2-3}

&Na\"{\i}ve&\textcolor{red}{Stomach, endoscopic mucosal resection; Gastrointestinal stromal tumor} \\\cline{2-3}

& DER $\mathcal{B}$=10 &\textcolor{blue}{ Urinary bladder, transurethral resection;} \textcolor{red}{1. No tumor present 2. Chronic granulomatous inflammation with foreign body reaction Note) The specimen includes muscle proper.}\\\cline{2-3}
 
& DER $\mathcal{B}$=50
&\textcolor{blue}{ Urinary bladder, transurethral resection; Non-invasive papillary urothelial carcinoma, low grade} \textcolor{red}{ Note) The specimen includes muscle proper.}\\\cline{2-3}
&Ours& \textcolor{blue}{Urinary bladder, transurethral resection; Non-invasive papillary urothelial carcinoma, low grade} \textcolor{red}{ Note) The specimen includes muscle proper.}\\\thickhline

\end{tabular}

\label{tab:qualitative}
\end{table}
%%%%%%%%%%%%%%%%%%%%%%%%%%%%%%%%%%%%%%%%%%%

%% file: 6_ablation.tex
\begin{table*}[!t]
\centering
\tiny
\caption{Ablation study. $\text{FR}$: footprint-based generative replay, $\text{RS}$: Report-style. }
 \label{tab:abl1}

\begin{tabular}{l|ccc|ccc|ccc|ccc}
\thickhline

 \begin{tabular}[c]{@{}c@{}}Exp.$\rightarrow$  \end{tabular}
& \multicolumn{3}{c|}{OS-R }
& \multicolumn{3}{c|}{OS-P }
& \multicolumn{3}{c|}{HS-C }
& \multicolumn{3}{c}{HS-D  }\\\hline

Method&  AVG $\uparrow$ &ILM $\uparrow$& BWT $\uparrow$
&  AVG $\uparrow$ &ILM $\uparrow$& BWT $\uparrow$
&  AVG $\uparrow$ &ILM $\uparrow$& BWT $\uparrow$
&  AVG $\uparrow$ &ILM $\uparrow$& BWT $\uparrow$\\\hline

na\"{\i}ve 
& 0.3535 & 0.4065 & -0.5086 
& 0.3428 & 0.3423 & -0.0308  
&  0.2949 & 0.3545 & -0.4239 
&0.2974& 0.3505& -0.4376
\\ \hline
%=======================%
\begin{tabular}[c]{@{}c@{}}Our \ding{55}$\text{FR}$ \ding{51}$\text{RS}$\end{tabular}
&0.3918& 0.4192& -0.4915
&0.3451&0.3447& -0.0251
& 0.2953& 0.3625& -0.4221
& 0.2985& 0.3507& -0.4412
\\

\begin{tabular}[c]{@{}c@{}}Our \ding{51}$\text{FR}$ \ding{55}$\text{RS}$\end{tabular}
& 0.6661& 0.7087& -0.0930
& 0.3427 & 0.3489 & -0.0155  
&  0.5075& 0.5990& \textbf{-0.0838}
&  0.5340&0.6300& -0.0577
\\

% &&\begin{tabular}[c]{@{}c@{}}Ours (\ding{51} style) (4,1)\end{tabular}
\begin{tabular}[c]{@{}c@{}}Our \ding{51}$\text{FR}$ \ding{51}$\text{RS}$\end{tabular}
& \textbf{0.6961}& \textbf{0.7319}&\textbf{ -0.0686}
&  \textbf{0.3517}& \textbf{0.3544}&\textbf{-0.0130}
&\textbf{0.5176}&\textbf{ 0.6094}& -0.0850
& \textbf{0.5473}&\textbf{ 0.6424}&\textbf{ -0.0413}
\\

\thickhline

\end{tabular}%
\end{table*}
%===============================================================

\subsection{Ablation}
Table~\ref{tab:abl1} confirms that footprint-based generative replay (FR) is a key contributor: removing FR causes a large drop in AVG/ILM and more negative BWT across settings. The report-style prototype alone yields only a marginal improvement over na\"{\i}ve sequential fine-tuning, indicating a limited but consistent benefit in the continual setup. Adding style conditioning on top of generative replay yields consistent gains in AVG/ILM (OS-R: 4.5/3.3\%, OS-P: 2.6/1.6\%, HS-C: 2.0/1.7\%, HS-D: 2.5/2.0\%) and improves retention on most shifts by reducing BWT (OS-R: 26.2\%, OS-P: 16.1\%, HS-D: 28.4\%), suggesting that domain-appropriate style prototypes, selected via domain-agnostic matching at inference, help preserve reporting conventions and mitigate CF under distribution shifts.

%% file: 7_Conclusions.tex
\section{Conclusion}
We presented an exemplar-free CL framework for WSI-to-report generation. The method summarizes each domain with compact visual footprints in a frozen embedding space, enabling replay via synthesized pseudo-WSIs and pseudo reports generated by an immediate teacher snapshot. We further incorporate per-domain report-style prototypes and footprint-matching strategy to support domain-agnostic style selection at inference. This framework enables continual adaptation under image and reporting-convention shifts without storing any WSI-report pairs.